\documentclass[9pt,journal]{IEEEtran}
\usepackage{amsmath,amssymb,amsfonts}
\usepackage{algorithmic}
\usepackage{textcomp}

\usepackage{array}
\usepackage{url}
\usepackage{color}
\usepackage{epsfig}
\usepackage{multirow}
\usepackage{float}
\usepackage{caption}
\usepackage{url,here}
\usepackage{dblfloatfix}
\usepackage{cite}
\usepackage{caption}
\usepackage{subcaption}
\usepackage{graphicx}
\def\BibTeX{{\rm B\kern-.05em{\sc i\kern-.025em b}\kern-.08em
    T\kern-.1667em\lower.7ex\hbox{E}\kern-.125emX}}
\begin{document}
	
\title{\LARGE \bf Efficacy of Haptic Pedal Feel Compensation\\ on Driving with Regenerative Braking}

\author{\centering Umut~Caliskan \hspace{55mm}        Volkan~Patoglu
\IEEEcompsocitemizethanks{\IEEEcompsocthanksitem U. Caliskan and V. Patoglu are with the Faculty of Engineering and Natural Sciences, Sabanci University, Istanbul, 34956, Turkey.\protect\\
E-mail: \{caliskanumut,vpatoglu\}@sabanciuniv.edu}
\thanks{Manuscript received Sep 27, 2019.}}

\markboth{IEEE Transactions on Haptics,~Vol.~X, No.~X, December~2019}%
{Shell \MakeLowercase{\textit{et al.}}: Bare Demo of IEEEtran.cls for Computer Society Journals}
%

\maketitle

\begin{abstract}
We study the efficacy of haptic pedal feel compensation on driving safety and performance during regenerative braking. In particular, we evaluate the effectiveness of the preservation of the natural brake pedal feel under two-pedal cooperative braking and one-pedal driving scenarios, through human subject experiments in a simulated vehicle pursuit task. The experimental results indicate that pedal feel compensation can significantly decrease the hard braking instances, improving safety for both two-pedal cooperative braking and one-pedal driving. Volunteers also strongly prefer compensation, while they equally prefer and can effectively utilize both two-pedal and one-pedal driving conditions. Furthermore, the beneficial effects of haptic pedal feel compensation is larger for the two-pedal cooperative braking case.
\end{abstract}



\vspace{-3mm}
\section{Introduction}
\label{sec:introduction}

With the current emphasis on decreasing smog forming emissions, electric and hybrid vehicles are becoming ubiquitous. The electric motors on these vehicles assume dual purpose. Not only can they be used to accelerate the vehicle, but they can also be employed as generators to decelerate the vehicle. The use of  electric motor for deceleration, by converting the kinetic energy of the vehicle into electrical energy to be stored in the battery, is called regenerative braking. Regenerative braking is crucial as it can significantly improve the range of the vehicle by improving its energy efficiency. Along these lines, it is desirable to employ regenerative braking as much as possible, while decelerating the vehicle.

During regenerative braking, the deceleration demand is measured based on a pedal displacement (as a signal) and appropriate resistance forces are applied to the vehicle through the electric motor to provide the desired deceleration level. However, there exists certain limitations of regenerative braking. The regenerative braking force depends non-linearly on the the speed of the vehicle, the state of the electrical motor and the charge level of the battery pack, at any given instant. Furthermore, regenerative braking can neither be applied at low speeds since sufficient braking forces cannot be generated, nor at high speeds since high voltages generated at these speeds may cause permanent damage to the battery. As a result, recruitment of  conventional friction brakes along side with regenerative braking is necessary to ensure safe deceleration at any speed~\cite{Wang2008}.

Conventional friction brakes are  commonly implemented using (electro)hydraulics. When the brake pedal is pressed, hydraulic fluid is pushed into the master cylinder where the hydraulic forces are multiplied by a brake booster and send to the activate the brake pads. Consequently, the brake pads apply longitudinal forces to the discs to create friction between the discs and the brake pads. Thanks to the hydraulic fluid, there exists a physical power exchange between the brake pedal and the friction brakes, and whenever a driver pushes the pedal, she/he feels the reaction forces due to this physical coupling. While brake-by-wire systems can be employed to remove this physical coupling to improve reaction times and reduce overall complexity of the braking system, current vehicle safety regulations do not allow for the complete removal of the physical connection.

Cooperative braking and one-pedal driving are the most commonly used approaches to blend regenerative and friction braking. In cooperative braking, when the brake pedal is pressed, the regenerative braking is utilized as much as possible to provide the demanded deceleration, while simultaneously charging the battery pack. The friction brakes are activated minimally, only to supplement regenerative braking, when the deceleration demand is higher than what can be provided solely by the regenerative braking~\cite{Kumar2015, Lindavaisoos2013}. In the literature, it has been shown that cooperative braking can be very efficient and recover up to 50\% more energy compared to alternative regenerative braking approaches~\cite{Kumar2015, Ohkubo2013}.

In one-pedal driving, the regenerative braking is controlled (partially) based on the accelerator pedal, such that when the driver releases the accelerator pedal, a certain amount of  regenerative braking is actived~\cite{Boekel2015,Saito2019}. One pedal driving promises ro reduce the reaction time of the drivers under braking~\cite{Nilsson2002} and simplify the overall driving experience~\cite{wen2018}. However, during one-pedal driving, the deceleration rate due to regenerative braking is limited either by the instantaneous regeneration capacity of the vehicle or by a pre-determined reasonable deceleration level selected to ensure the driver comfort. Furthermore, an emergency brake pedal physically coupled to the friction brakes is still needed such that the driver can intervene for higher deceleration rates during emergency braking situations.

In both two-pedal cooperative braking or one-pedal driving, when the regenerative and friction brakes are simultaneously activated by the driver interacting with the (emergency) brake pedal, the conventional haptic brake pedal feel is disturbed due to regenerative braking. In particular, while there exists a physical coupling between the brake pedal and the conventional (electro)hydraulic friction brakes, no such physical coupling exists for the regenerative braking.  As a result, no reaction forces are fed back to the brake pedal, resulting in a unilateral power flow between the driver and the vehicle. Consequently, the relationship between the brake pedal force and the vehicle deceleration is strongly influenced by the regenerative braking. When regenerative/friction braking is activated/deactivated, the pedal response may change abruptly, resulting in rapid softening/stiffening of the brake pedal.  This unfamiliar response of the brake pedal poses a safety concern, since it may negatively impact the driver performance.

Reaction forces due to regenerative braking can be fed back to the brake pedal, through actuated pedals that re-establish the bilateral power flow between the brake pedal and the vehicle to recover the natural haptic  pedal feel. Along these lines, electro-hydraulic~\cite{Kumar2015,Ohkubo2013,Zhang2012} 
and electro-mechanical~\cite{Stachowski2004,Derasorio2010} force-feedback brake pedals have been proposed in the literature. The authors have also proposed a force-feedback brake pedal with series elastic actuation (SEA) to preserve the conventional brake pedal feel during cooperative regenerative braking~\cite{IROS2018}.

In this paper, a human subject experiment is conducted to test the efficacy of haptic pedal feel compensation on driving performance during cooperative braking. In a simulated vehicle pursuit scenario, a torque-controlled dynamometer is utilized for rendering the reaction forces due to friction braking, while an SEA brake pedal is employed to compensate for the disturbing effects of regenerative braking, to recover a natural brake pedal feel. The driving performance of volunteers under  regenerative braking with and without haptic pedal feel compensation, under one-pedal driving and two-pedal cooperative braking conditions are reported. This work significantly extends the preliminary user study presented in~\cite{IROS2018} by the addition of the dynamometer to render reaction forces due to the friction brake, the utilization of an accelerator pedal, the addition of the one-pedal driving condition and more extensive evaluations based on a new experimental protocol and multiple performance metrics.

\begin{figure}[b]
	\centering
\vspace{-.5\baselineskip}
	\resizebox{.70\columnwidth}{!}
	{\rotatebox{0}{\includegraphics[width=1\textwidth]{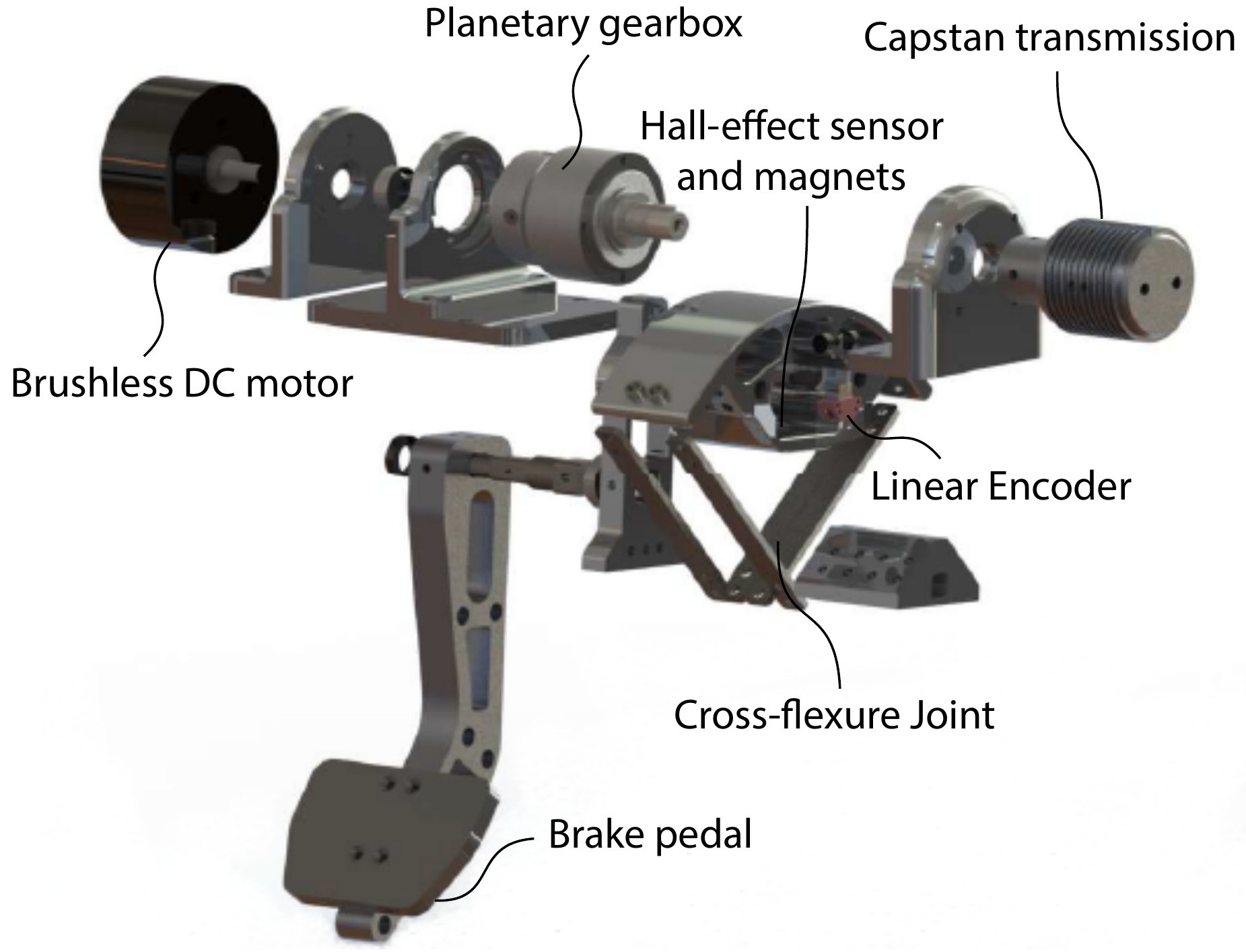}}}
	\vspace{-.5\baselineskip}
	\caption{Mechatronic design of the SEA brake pedal and the dynamometer} \label{fig:explodview}
\end{figure}

\section{Experimental Setup}
\label{sec:Experimental Setup}

\subsection{Series Elastic Brake Pedal}

Figure~\ref{fig:explodview} presents an exploded view of the series elastic brake pedal, whose initial design has been detailed in~\cite{IROS2018}. The device is actuated by a brushless DC motor equipped with an optical encoder to provide 1~Nm continuous torque output. A low-friction planetary gear train with 10:1 reduction is followed by a  capstan transmission with 4:1 reduction, to amplify the torque output of the DC motor. The sector pulley of the capstan transmission is attached to brake pedal through a leaf spring based cross-flexure pivot that serves as a robust and simple compliant joint with a large deflection range. A Hall-effect sensor and a linear encoder are used for measuring the deflection of the cross-flexure pivot and estimating the interaction torques at the pedal.
Once the interaction forces are estimated, closed-loop force control is implemented.

Thanks to its series elasticity, the force-feedback brake pedal can utilize robust controllers to achieve high fidelity force control, possesses favourable output impedance characteristics over the entire frequency spectrum, and can be implemented in a relatively compact package using low-cost components.

The SEA brake pedal used in this study features a force control bandwidth of 13~Hz for forces up to 75~N and can continually provide pedal forces over 200~N to the driver's foot.

\begin{figure}[t]
	\centering
	\resizebox{0.75\columnwidth}{!}
	{\rotatebox{0}{\includegraphics[width=1\textwidth]{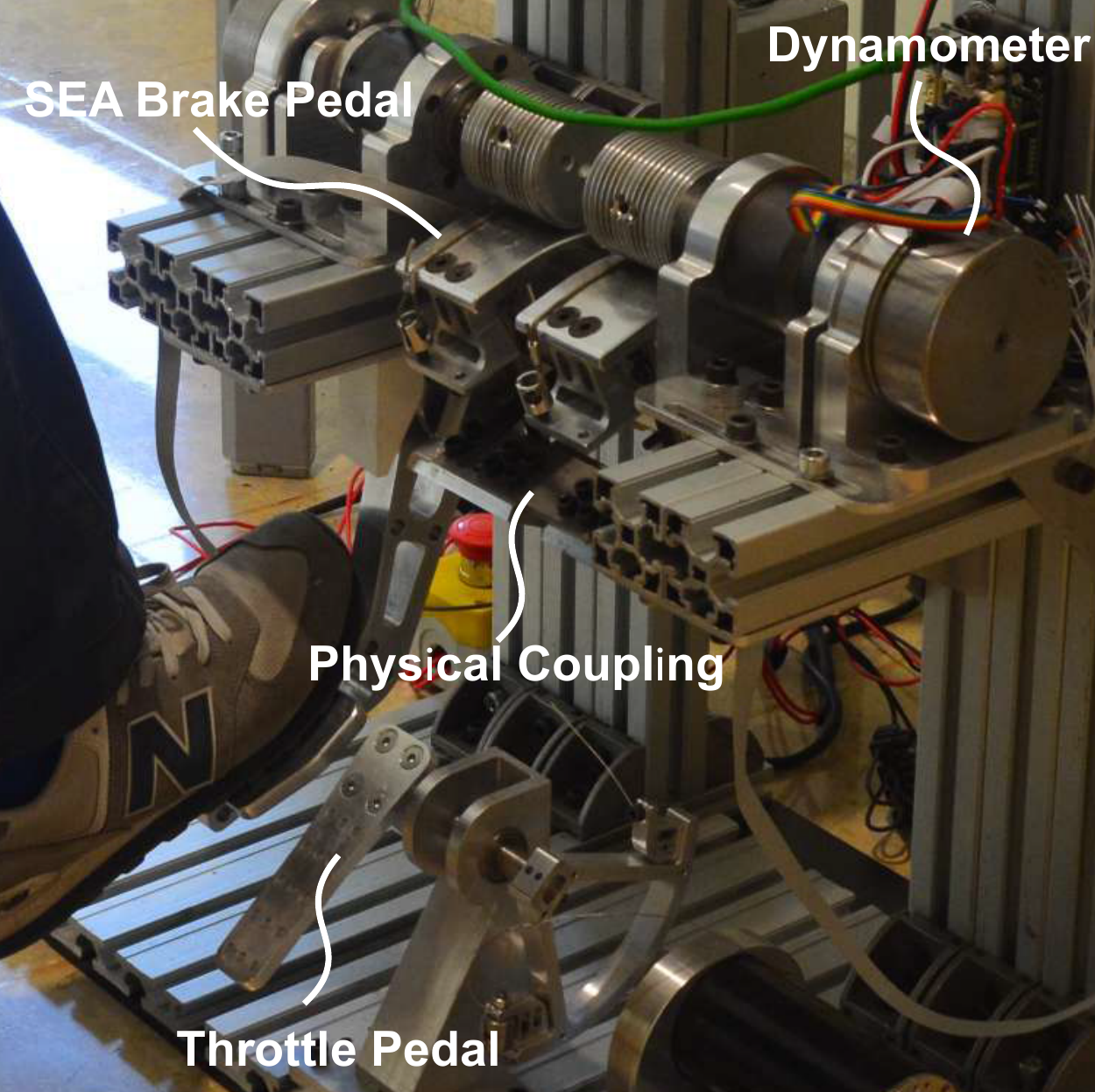}}}
    \vspace{-.25\baselineskip}
	\caption{Haptic pedal feel rendering platform for cooperative braking} \label{fig:experimentalsetup}
    \vspace{-.75\baselineskip}
\end{figure}

\subsection{Haptic Pedal Feel Rendering Platform}

Figure~\ref{fig:experimentalsetup} presents a solid model of the haptic pedal feel rendering platform developed for testing different regenerative braking approaches. The system consists of a SEA brake pedal and a torque controlled dynamometer that share identical designs, as depicted in~Figure~\ref{fig:explodview}. The two force feedback devices are mechanically coupled to each other through a rigid connection.

The dynamometer is used to render (electro)hydraulic friction brake reaction forces originating from the vehicle’s controllable master cylinder, as well as other forces/disturbances acting on the brake pedal, while the force-feedback pedal is used to implement cooperative braking algorithms to compensate for the disturbance effects and to recover the natural brake pedal feel.

Furthermore, to enable simulation of one-pedal driving, an open-loop impedance controlled throttle pedal is included to the system as presented in Figure~\ref{fig:experimentalsetup}. The throttle pedal consists of a direct drive motor with a 10:1 ratio capstan  transmission such that forces up to 75~N can be provided the driver's foot.

\begin{figure*}[t]
	\centering \resizebox{2.05\columnwidth}{!}{\rotatebox{0}{\includegraphics{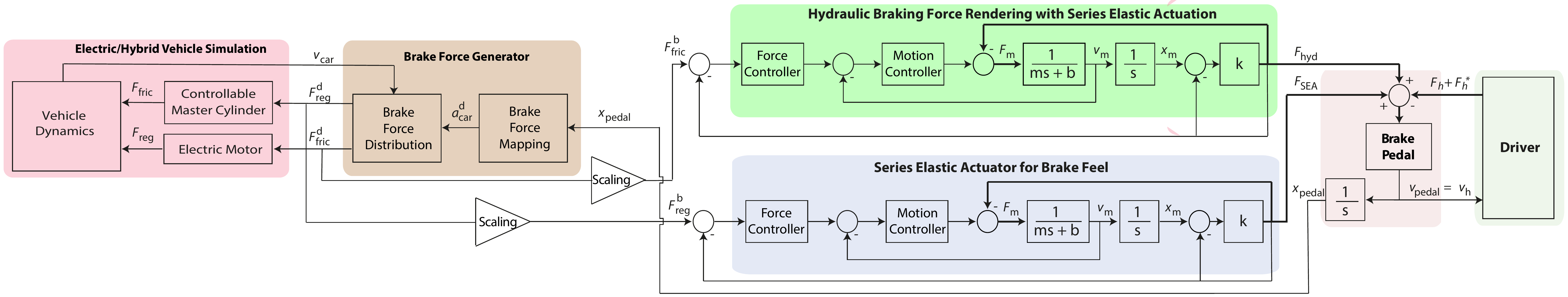}}}
	\vspace{-1.65\baselineskip}
	\caption{Control block diagram of haptic pedal feel rendering platform}
	\vspace{-\baselineskip}
	\label{Fig:block_diag}
\end{figure*}

\subsection{Control of the Haptic Pedal Feel Rendering Platform}

Figure~\ref{Fig:block_diag} presents the block diagram used to control the haptic pedal feel rendering platform. In the figure, the thick lines denote power coupling and the thin lines represent signals. Symbols $m$ and $b$ denote the effective inertia and damping of the identical SEA devices. Human applied forces are indicated by two distinct components: $F_h$ representing the passive component and $F_h^*$ denoting the intentionally applied active component, which are assumed to be independent of the system states, such that coupled stability can be concluded through the frequency domain passivity framework~\cite{Colgate1988}.

In Figure~\ref{Fig:block_diag}, after appropriate scaling, the regenerative brake force demand $F^d_{reg}$ is passed to the SEA brake pedal as a reference force. The SEA pedal relies on closed-loop force control to ensure that this reference force is rendered to the driver with high fidelity. Similarly, the friction force brake demand $F^d_{fric}$ is passed to the dynamometer as a reference force such that (electro)hydraulic friction brake reaction forces originating from the vehicle’s controllable master cylinder are rendered to the driver. Consequently, the driver feels the force feedback from the total braking force applied to the vehicle, that is, the sum of forces from the friction brakes $F_{hyd}$ through the dynamometer and forces from the regenerative brakes $F_{SEA}$ through the SEA brake pedal.

The force/torque control of the brake pedal and the dynamometry are implemented as independent control loops, such that they can be run at different control rates and in an unsynchronized manner to be able to render more realistic disturbance and compensation forces. Independent real-time cascaded PI controllers are implemented for the control of series elastic actuators. In this cascaded controller, the fast inner-loop running at 2.5~kHz controls the velocity of the geared motor, rendering it into an ideal motion source by compensating for imperfections in the power transmission, such as friction and stiction in the gearbox. The outer-loop, implemented at 1~kHz, controls the interaction torque based on the deflection feedback from the compliant element. The coupled stability of the cascaded control architecture of SEA is guaranteed within the frequency domain passivity framework with the proper choice of controller gains, as detailed in~\cite{FatihEmre2019}.

\section{Haptic Pedal Feel Compensation}
\label{sec:Compensation}

In this section, pedal feeling rendering algorithms for two-pedal cooperative braking and one-pedal driving are detailed.

\vspace{-1mm}
\subsection{Conventional Haptic Brake Pedal Feel}
\label{sec:pedalfeel}
The conventional haptic brake pedal feel to be recovered under the intervention of regenerative braking is mathematically modeled as
\small
 \[ F_{pedal}~[\text{N}] = \begin{cases}
0.80~x_{pedal}+18.17 & \hspace{12mm}  x_{pedal} \leq 20~\text{mm}  \\
3.92~x_{pedal}-44.23 & 20~\text{mm} < x_{pedal} \leq 80~\text{mm} \\
   \end{cases}
\] \normalsize
\noindent where $x_{pedal}$ denotes the pedal displacement with a maximum stroke of 80~mm and $F_{pedal}$ is the total  pedal force~\cite{Spandoni2013}.

\begin{figure*}
	\centering
	\begin{subfigure}[b]{0.45\textwidth}
		\centering
		\includegraphics[width=\textwidth]{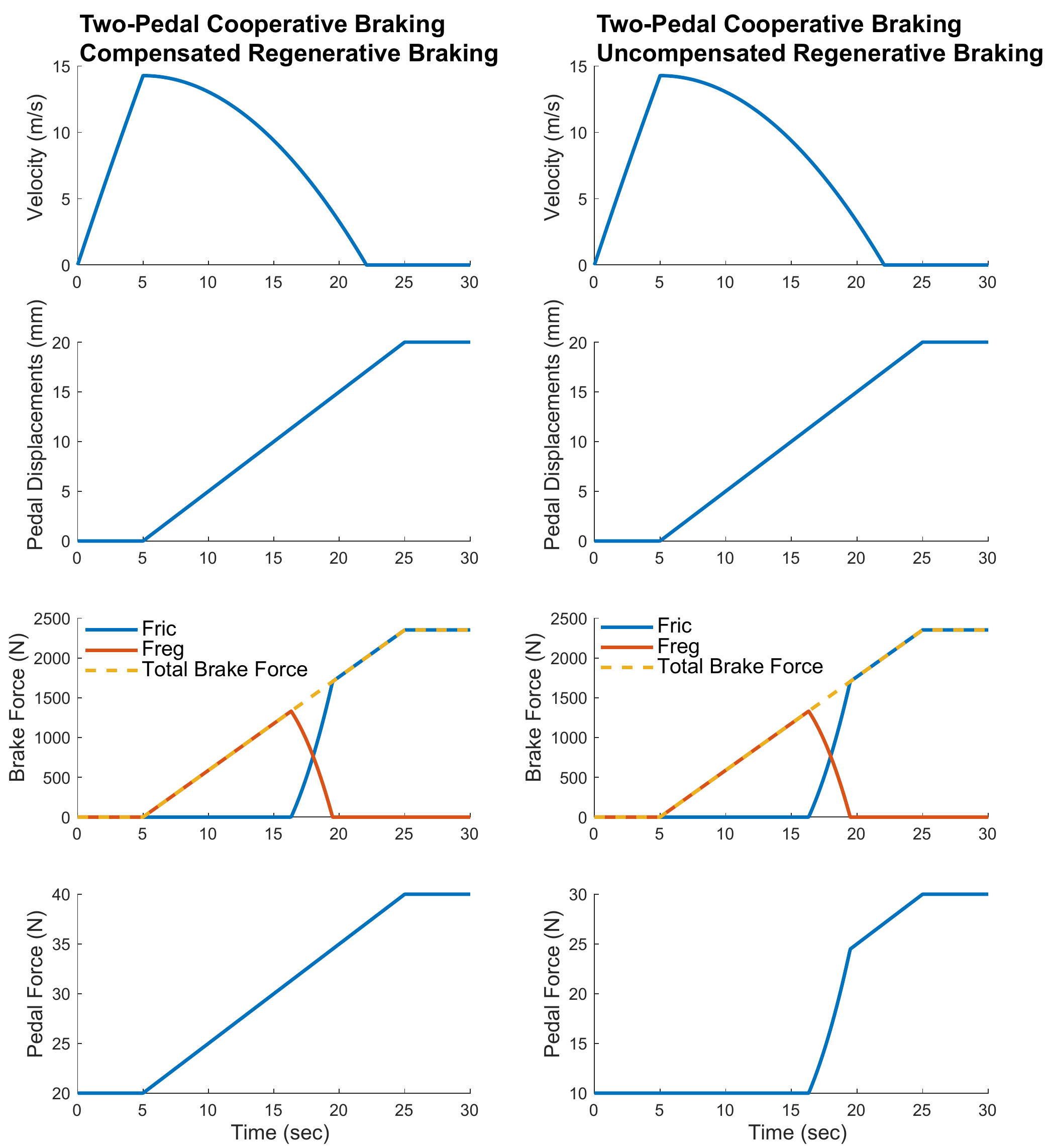}
        \vspace{-1.5\baselineskip}
		\caption{A sample scenario for two-pedal cooperative braking with and without haptic pedal feel compensation}
		\label{fig:twopedalcond}
	\end{subfigure}
	\hfill
	\begin{subfigure}[b]{0.45\textwidth}
		\centering
		\includegraphics[width=\textwidth]{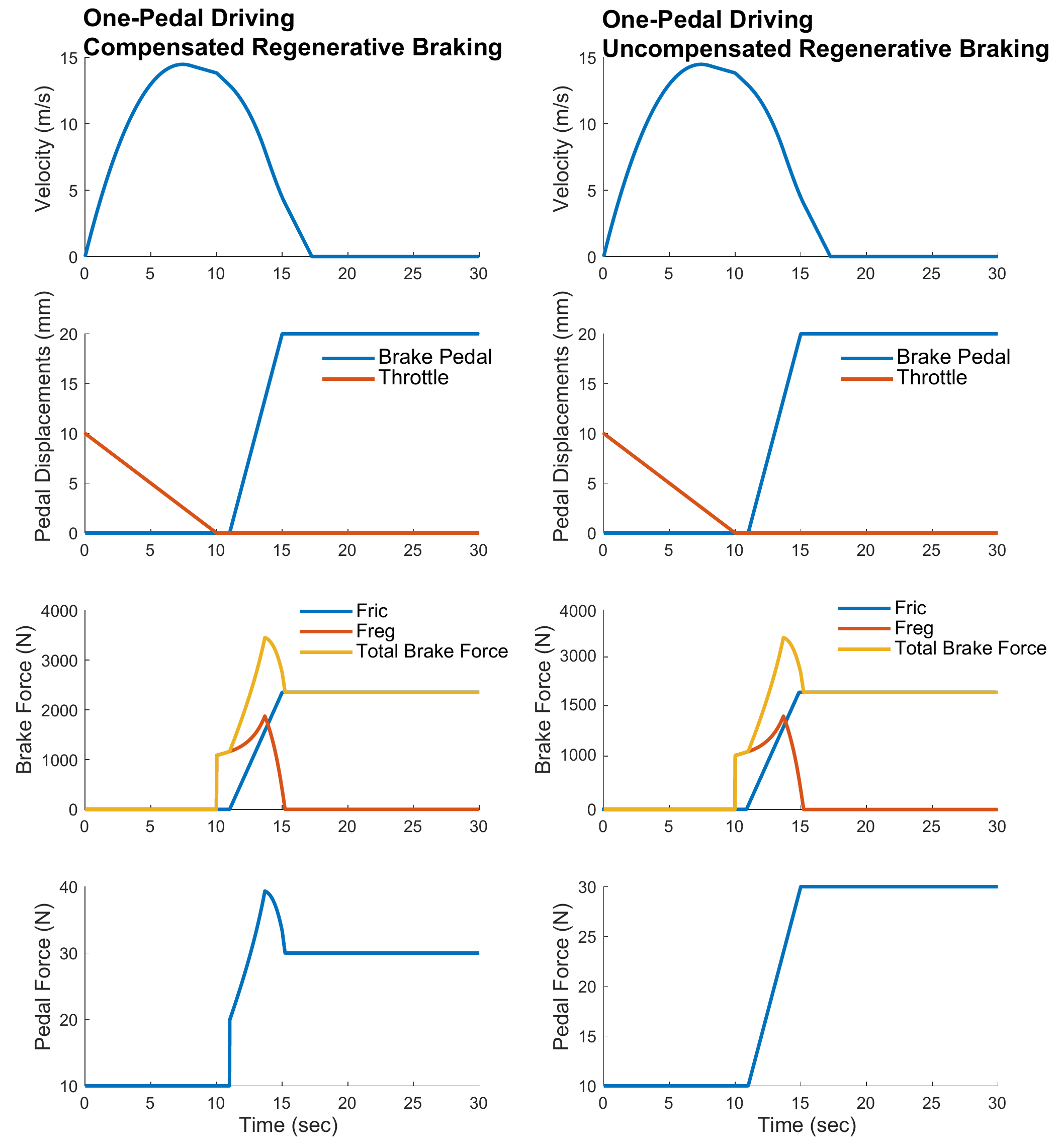}
        \vspace{-1.5\baselineskip}
		\caption{A sample scenario for one-pedal driving with and without haptic pedal feel compensation}
		\label{fig:onepedalcond}
	\end{subfigure}
    \vspace{-.5\baselineskip}
    \caption{Vehicle states for four different driving conditions}
    \vspace{-\baselineskip}
	\label{fig:compvsone}
\end{figure*}

\vspace{-1mm}
\subsection{Brake Pedal Displacement to Deceleration Mapping}

The brake pedal displacement $x_{pedal}$  is mapped to the deceleration demand $a_{car}^d$  according to the following function as proposed in~\cite{Spandoni2013}.
\small
 \[ a_{car}^d~[\text{m/sec}^2] = \begin{cases}
      -(0.01~x_{pedal})g  & \hspace{12mm} x_{pedal}\leq 20~\text{mm} \\
      -(0.02~x_{pedal}-0.2)g  & 20~\text{mm} \leq x_{pedal} \leq 80~\text{mm} \\
   \end{cases}
\] \normalsize
\noindent where $g$ represents the gravitational acceleration.

\vspace{-1mm}
\subsection{Brake Pedal Force due to Friction Braking}
\label{sec:pedalforcemap}

To render the reaction forces on the brake pedal during friction braking, the brake pedal position is mapped to the dynamometer torque as
\small
\[ F_{Fric}^b~[\text{N}] = \begin{cases}
0.16~x_{pedal}+3.63 & \hspace{12mm}  x_{pedal} \leq 20~\text{mm}  \\
0.78~x_{pedal}-8.84 & 20~\text{mm} < x_{pedal} \leq 80~\text{mm} \\
\end{cases}
\] \normalsize
\noindent where $F_{Fric}^b$ denotes the hydraulic friction braking forces applied by the dynamometer.

\vspace{-1mm}
\subsection{Brake Force Distribution}
\label{sec:brakeforce}

Brake force distribution is decided based on the deceleration demand~$a_{car}^d$ from the driver, instantaneous vehicle speed $v_{car}$, battery charge level and the road conditions.  A simple model of instantaneous regenerative braking force is employed as $F_{reg} = \frac{P_m}{v_{car}}$, where $P_m$ denotes the constant braking power of the electric motor, and $v_{car}$ is the instantaneous velocity of the vehicle~\cite{Kumar2015}. Note that regenerative braking forces $F_{reg}$ cannot be generated below/above some critical speed, in particular, below 4~m/sec (15~km/h) and above 33~m/sec (120~km/h). To avoid inducing any sudden changes in regenerative braking force, linear interpolation is used around the critical speeds to smooth out the transition. The regenerative braking force to pedal force mapping is given as follows.
 \small
 \[ F_{reg}^b~[\text{N}] = \begin{cases}
 0.0164~F_{reg}-44.21 & \hspace{8mm} F_{reg}\geq 2352~\text{N} \\
 0.0068~F_{reg}+18.17  & 0~\text{N} \leq F_{reg} < 2352~\text{N} \\
 \end{cases}
 \] \normalsize

Given the regenerative braking capacity at any instant and neglecting the road conditions for simplicity, the brake force distribution block determines the amount of regenerative and friction braking that needs to be employed, based on the one-pedal versus two-pedal driving condition.

Sample cooperative braking scenarios with and without haptic brake pedal feel compensation are presented for two-pedal and one-pedal driving in Figures~\ref{fig:compvsone}(a) and~\ref{fig:compvsone}(b), respectively. In the first row of the figures, the velocity of the vehicle is depicted, while the pedal displacement is presented in the second row. For the one-pedal driving case, the throttle displacement is also presented. In the third row, the regenerative braking forces, friction brake forces and total brake forces are depicted. The last row presents the pedal forces felt by the driver. In these sample scenarios, pedals are assumed to be displaced in a linear manner, for the simplicity of presentation.

\subsubsection{Two-Pedal Cooperative Braking}

In two-pedal cooperative braking, regenerative brake is activated by pressing the brake pedal. When there is a deceleration demand from the driver, the regenerative braking is utilized as much as possible. When the deceleration demand is higher than that can be supplied by the regenerative braking, the friction brake is activated. In the uncompensated case, there exists no pedal force due to regenerative braking, while in the compensated case, relevant pedal forces are rendered to the pedal as discussed in previous subsection.

In Figure~\ref{fig:compvsone}(a), when the driver presses the brake pedal at t~=~5~s, regenerative brake is employed to the maximum capacity.  The regenerative braking forces increase in a nonlinear fashion, as the vehicle slows down. Note that no pedal force exists for the non-compensated case when friction brake is not in use.
Since the regenerative braking forces cannot be generated at velocities lower than 4~m/sec, the friction brake is employed at t = 16~s such that the desired deceleration demand can be delivered. Starting this instant, brake pedal forces go through a sharp increase in the uncompensated condition until the friction brake takes over the whole braking at t = 20~s. After t = 20~s, the uncompensated pedal feels like a conventional friction brake. Note that the compensation eliminates the discontinuities and stiffening/softening of haptic pedal feel due to regenerative braking and delivers a continuous conventional brake pedal forces throughout the cooperative braking.

\subsubsection{One-Pedal Driving}

One-pedal driving and two-pedal cooperative braking differ in that regenerative braking is activated when the throttle pedal is released in one-pedal driving. In particular, when the driver releases the throttle pedal, the maximum available regenerative braking force is utilized until a threshold (chosen as 0.32g in this study) after which the force is saturated not to induce an uncomfortable deceleration level. If the driver presses the emergency brake pedal, further use of regenerative braking may be activated as in cooperative braking, while typically the friction brake is activated, as most capacity of regenerative braking is already in use. In the uncompensated case, there exists no pedal force due to regenerative braking, while in the compensated case, relevant pedal forces are rendered to the emergency brake pedal to achieve a linear relationship with the total braking force.

In Figure~\ref{fig:compvsone}(b), the driver releases the throttle pedal at t = 10~s, which activates the regenerative braking, but does not render any forces to the emergency brake pedal in both cases, as it is not being pushed yet. The displacement of the emergency brake pedal is increased linearly during t = 11--15~s and the friction brake is activated, as the deceleration from regenerative braking is not sufficient to provide the demanded deceleration. In the uncompensated case, the driver feels only the reaction forces from the friction brake. While this force is continuous, the mapping between the pedal force and the total brake force is nonlinear. In the compensated case, this mapping is linear.
%

\vspace{-4mm}
\section{User Evaluations}
\label{sec:User Evaluations}

\vspace{-2mm}
\subsection{Participants}

Ten volunteers (8 males and 2 female) with ages between 22 to 28 participated in the experiment. All participants had active driver's licenses and none of them had any prior experience with vehicles equipped with regenerative braking. All participants signed an informed consent approved by the IRB of Sabanci University.

\vspace{-2mm}
\subsection{Driving Simulator}

\begin{figure}[b!]
	\centering 	\vspace{-.5\baselineskip}
	\resizebox{0.7\columnwidth}{!}
	{\rotatebox{0}{\includegraphics[width=1\textwidth]{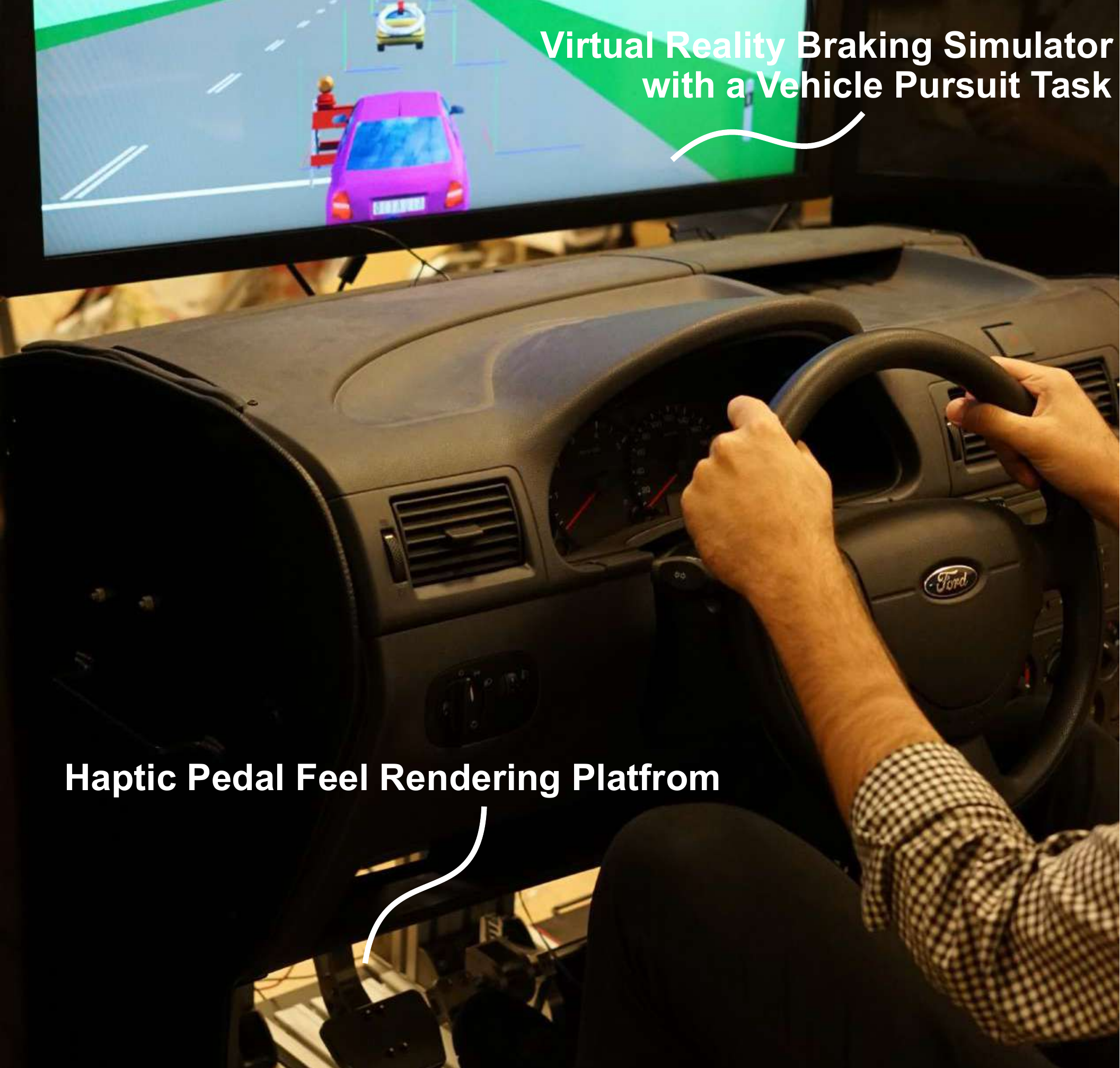}}}
	\vspace{-.5\baselineskip}
	\caption{Cooperative braking simulator} \label{fig:experimentalsetup2}
\end{figure}

The simulator setup consisted of an SEA brake pedal, a dynamometer, a throttle pedal and a vehicle simulator, as presented in Figure~\ref{fig:experimentalsetup2}. Participants were seated in a vehicle seat and adjusted the seat position according to their preferred driving position. The simulator provided visual feedback through two flat screens displays. The front screen displayed the simulated vehicle pursuit scenario, while the left monitor showed the vehicle speed.

\vspace{-1mm}
\subsection{Task}

The pursuit task is based on a simplified version of the Crash Avoidance Metrics Partnership~(CAMP) protocol~\cite{Hoffman2014}. The simulation took place on a virtual straight road of 1500~m, where the controlled vehicle followed a leading vehicle. The leading vehicle accelerated at 0.2g until it reached the target speed of 50~km/h. Once it reached 50~km/h the leading vehicle decelerated until stop, and then after waiting for a short random interval, it re-accelerated back to 50~km/h. In particular, the leading vehicle decelerated with 0.19g, 0.28g and 0.39g at random instances within the 0--500~m, 500~m--1000~m, and 1000~m--1500~m stretches of the road. The leading vehicle stopped permanently at the end of the road.

Initially, the following vehicle was placed 15~m behind the leading vehicle. The volunteers operated the throttle for acceleration and SEA brake pedal for (emergency) braking. The volunteers were asked to keep a 30~m distance to the lead car.

\begin{figure*}[b]
	\centering
    \vspace{-1.5\baselineskip}
	\begin{subfigure}[b]{0.34\textwidth}
		\centering
		\includegraphics[width=\textwidth]{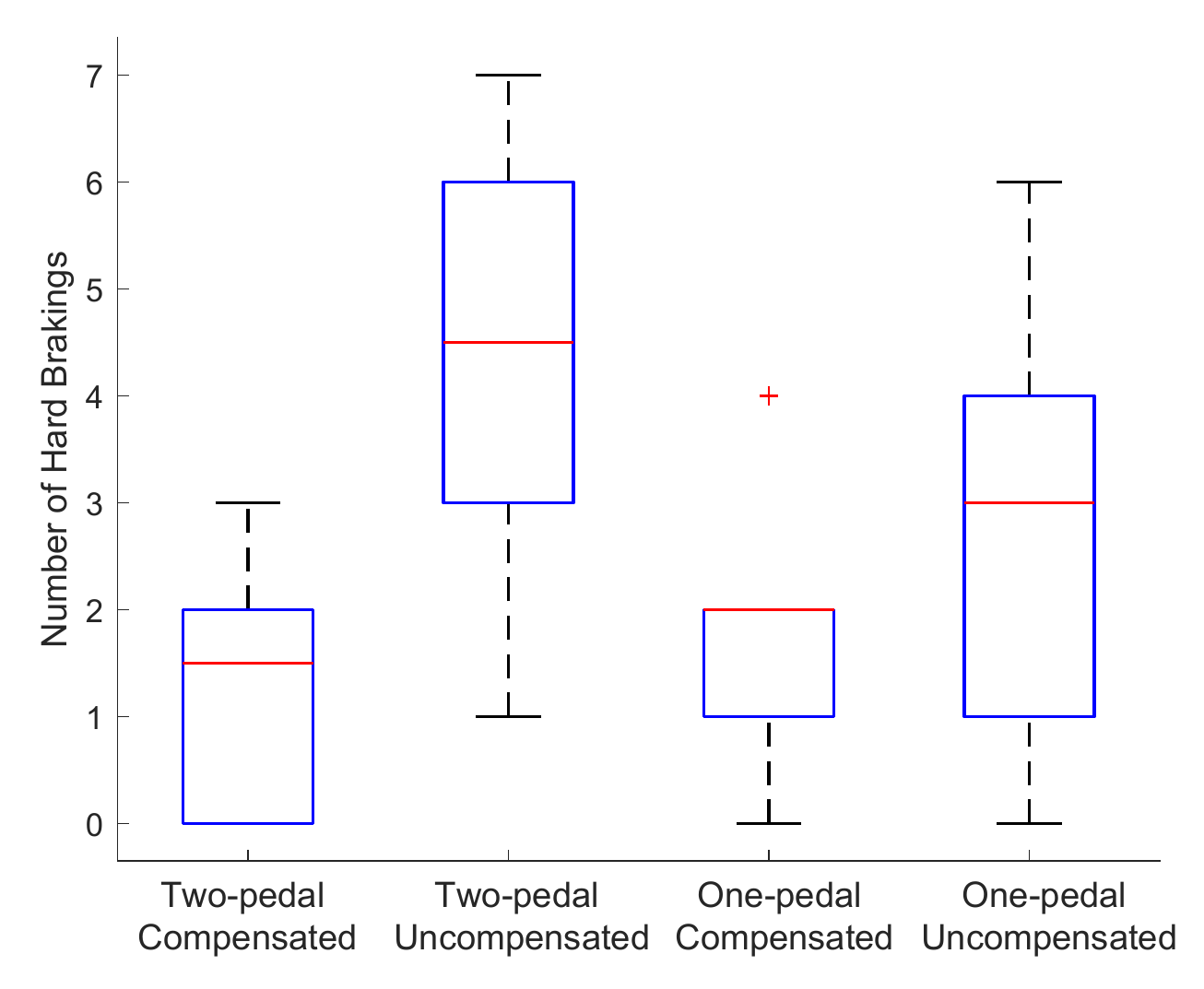}
        \vspace{-1.90\baselineskip}
		\caption{Box plot of number of hard brakings}
		\label{fig:numberofhard}
	\end{subfigure}
		\hfill
		\begin{subfigure}[b]{0.31\textwidth}
			\centering
			\includegraphics[width=\textwidth]{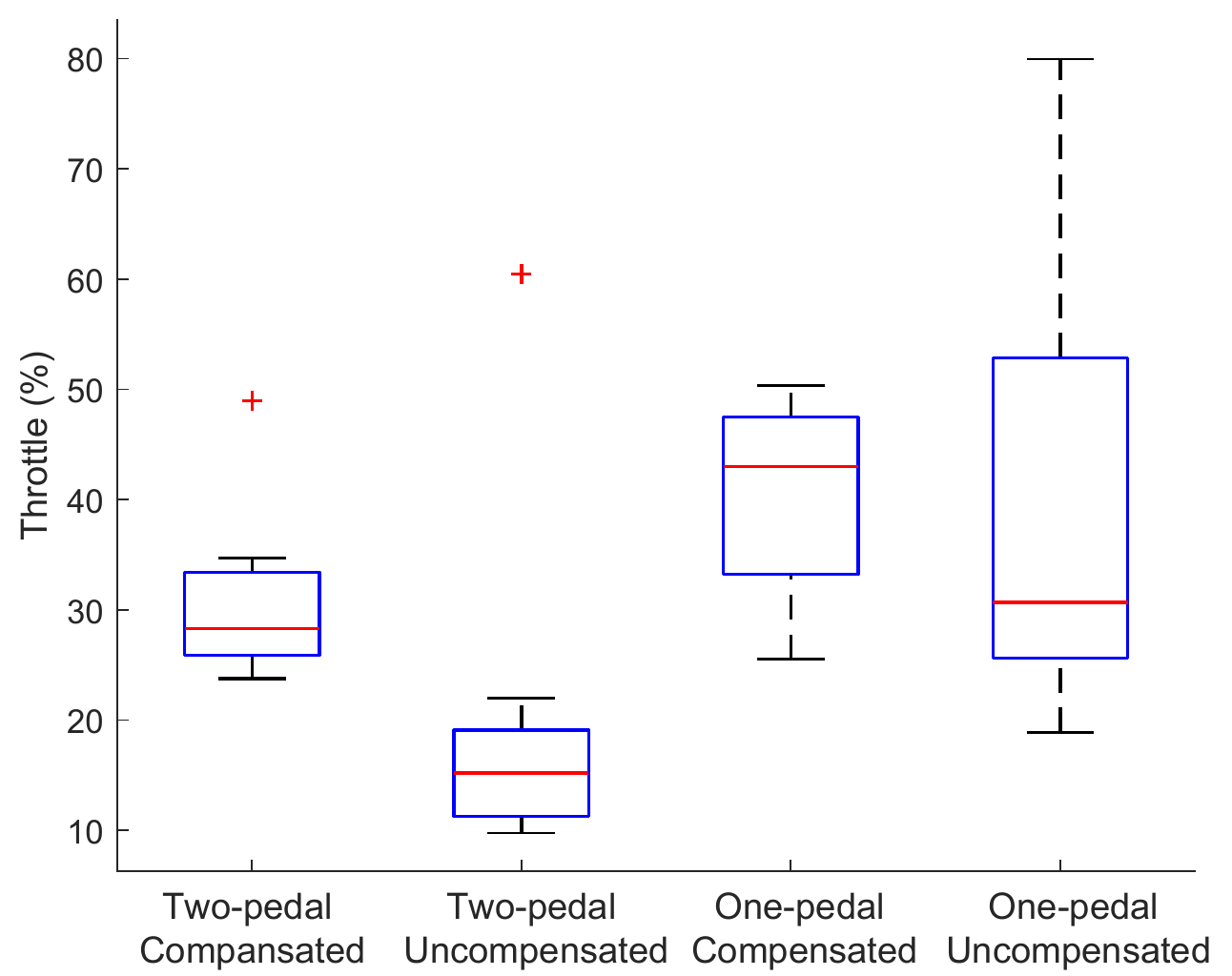}
            \vspace{-1.5\baselineskip}
			\caption{Box plot of percent throttle use}
			\label{fig:throttle}
		\end{subfigure}
			\hfill
		\begin{subfigure}[b]{0.32\textwidth}
			\centering
			\includegraphics[width=\textwidth]{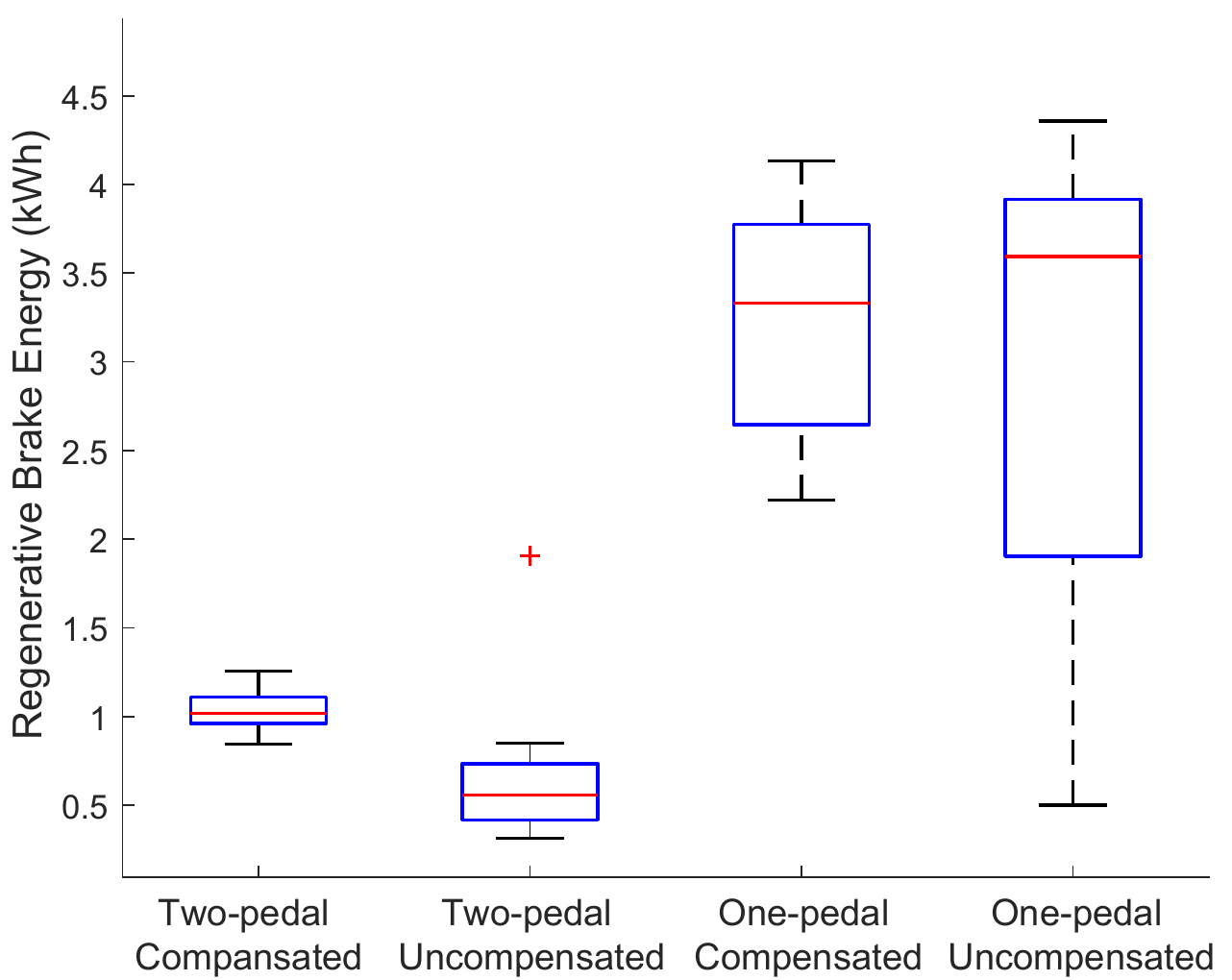}
            \vspace{-1.5\baselineskip}
			\caption{Box plot of regenerative braking energy}
			\label{fig:freg}
		\end{subfigure}
    \vspace{-.5\baselineskip}
	\caption{Summary statistics of the important performance metrics}
	\label{fig:boxplot}
	\vspace{-.25\baselineskip}
\end{figure*}

\vspace{-2mm}
\subsection{Experimental Procedure}

Effect of two main factors of compensation  and pedal type are investigated. In particular the within subjects experiment protocol involved \textit{two-pedal uncompensated}, \textit{two-pedal compensated}, \textit{one-pedal uncompensated} and \textit{one-pedal compensated} conditions tested on the same volunteers. At the beginning of experiments an unrecorded session was implemented, during which all four conditions were displayed to the volunteers in a randomized order to help them familiarize with the braking simulator. Then, volunteers were assigned to test conditions in a randomized order. The volunteers were informed about one pedal versus two pedal driving condition, but not about the existance/lack of compensation. After each trial, they were asked to recognize the existence of compensation.

\subsection{Performance Metrics}

Several quantitative metrics are defined to evaluate the driving performance of the participants.
The number of times \textit{hard brakings} were necessitated during the trials is selected as a performance metric, as large decelerations are potentially dangerous. In particular, decelerations over 0.5g are considered as hard braking~\cite{Lee2002}.

For driving performance analysis, the distance between two vehicles is selected as the performance metric. In particular, $\%\:RMSE$ is calculated with respect to the instructed distance of 30~m. 

To evaluate the energy efficiency of driving, regenerated energy of each session is calculated by adding the regenerative power at each time step. Furthermore, percent throttle use is also computed.

Finally, the volunteers are asked to fill in a short questionnaire to help evaluate their qualitative preferences among the test conditions. The questionnaire included nine questions as presented in Table~\ref{Tab:survey}. A 5-point Likert scale is used to indicate preferences, where 5 denotes strong agreement and 1 denotes strong disagreement.

\vspace{-3mm}
\subsection{Analysis}

Two-way repeated measures ANOVA  is conducted to determine the significant effects on the quantitative metrics. The within-within factors are taken as compensation (compensated/uncompensated) and pedal type (two-pedal/one-pedal). Box plots of important metrics are present to enable multi-comparisons and effect size evaluations.

\vspace{-2mm}
\section{Results}
\label{sec:Results}

\vspace{-1mm}
\subsection{Quantitative Metrics}

\subsubsection{Safety}
\label{sec:safety}

Figure~\ref{fig:boxplot}(a) presents the box plot for the number of hard brakings.
Two-way repeated measures ANOVA indicates that the interaction of compensation and pedal type factors is significant with  $F(1,9)=9.51, \: p=0.014$. The compensation is significant, while the pedal type is not significant at the  $p < 0.05$ level. 

For the simple main effect analysis, the data is first split for two-pedal and one-pedal driving conditions.
For the two-pedal driving condition,  hard brakings in the compensated case ($M\:=\:1.2,\:SD\:=\:0.33$) are significantly lower than the uncompensated case ($M\:=\:4.4,\:SD\:=\:0.56$) with $F(1,9)=39.05,\: p<0.001$.
The effect size is significant as the number of hard brakings have increased more than 3.5 times in the uncompensated case. Similarly, for the one-pedal driving condition, hard brakings in the compensated case ($M\:=\:1.8,\:SD\:=\:0.36$) are significantly lower than the  uncompensated case ($M\:=\:2.8,\:SD\:=\:0.53$) with $F(1,9)=5.63,\: p=0.042$. The effect size is also significant as the number of hard brakings has increased by 55\% in the uncompensated case.

The data is also split for compensated and uncompensated conditions. For the uncompensated condition,
hard brakings instances in the two-pedal condition ($M\:=\:4.4,\:SD\:=\:0.56$) are significantly higher than the one-pedal case ($M\:=\:2.8,\:SD\:=\:0.53$) with $F(1,9)=5.43,\: p=0.045$. The effect size is significant as the number of hard brakings has increased more than 57\% in the two-pedal case. For the compensated group, pedal type is not a significant factor at the  $p<0.05$ level.

\subsubsection{Driving Performance}

Two-way repeated measures ANOVA indicates that compensation, pedal type, and interaction are not significant factors for the $\%\:RMSE$ metric quantifying the tracking performance at the  $p<0.05$ level.

\subsubsection{Energy Efficiency}
\label{sec:energy}

Figure~\ref{fig:boxplot}(b) presents the box plot for the percent throttle use. Two-way repeated measures ANOVA indicates that one-pedal driving ($M\:=\:40.1,\:SD\:=\:3.1$) results in significantly higher throttle use compared to two-pedal cooperative braking ($M\:=\:25.86,\:SD\:=\:4.25$) with  $F(1,9)=6.92,\: p=0.034$. Compensation and interaction are not significant at the $p<0.05$ level. The effect size is significant as the throttle use has increased by 60\% in the one-pedal driving case.

Figure~\ref{fig:boxplot}(c) presents the box plot for the regenerated braking energy. One-pedal driving ($M\:=\:3.096,\:SD\:=\:0.25$) results in significantly higher regenerated energy compared to two-pedal cooperative braking ($M\:=\:1.035,\:SD\:=\:0.045$) with  $F(1,9)=70.15,\: p<0.001$, while compensation and interaction are not significant at the  $p<0.05$ level. The effect size is significant as 3 times more energy is regenerated during the one-pedal driving.

\begin{center} 
	\begin{table*}[t] \caption{Survey Questions and Summary Statistics} \vspace{-5mm}
		\label{Tab:survey} \centering
	    \resizebox{1.55\columnwidth}{!}{\begin{tabular}{m{5.75in}cc}
				\multicolumn{1}{m{5.75in}}{\vspace{1mm}\textbf{}\hspace{10mm} Cronbach $\alpha > $ 0.96 } & Mean & $\sigma^2$ \\ \hline \hline
				\multicolumn{1}{m{5.75in}}{\vspace{1mm}\textbf{Q1}: Do you feel the intervention of friction brake under the two-pedal uncompensated regenerative brake condition?} & 4.6 & 0.49\\ \hline
				\multicolumn{1}{m{5.75in}}{\vspace{1mm}\textbf{Q2}: Can you stop the car within the desired distance and time under the two-pedal uncompensated regenerative brake condition?} & 3.2 & 0.60\\ \hline
				\multicolumn{1}{m{5.75in}}{\vspace{1mm}\textbf{Q3}: Do you feel the intervention of friction brake under the two-pedal compensated regenerative brake condition?} & 1.4 & 0.49\\ \hline
				\multicolumn{1}{m{5.75in}}{\vspace{1mm}\textbf{Q4}: Can you stop the car within the desired distance and time under the two-pedal compensated regenerative brake condition?} & 4.2 & 0.60\\ \hline
				\multicolumn{1}{m{5.75in}}{\vspace{1mm}\textbf{Q5}: Do you feel the intervention of friction brake under the one-pedal uncompensated regenerative brake condition?} & 1.9 & 0.70\\ \hline
				\multicolumn{1}{m{5.75in}}{\vspace{1mm}\textbf{Q6}: Can you stop the car within the desired distance and time under the one-pedal uncompensated regenerative brake condition?} & 3.0 & 0.63\\ \hline
				\multicolumn{1}{m{5.75in}}{\vspace{1mm}\textbf{Q7}: Do you feel the intervention of friction brake under the one-pedal compensated regenerative brake condition?} & 2.1 & 0.70\\ \hline
				\multicolumn{1}{m{5.75in}}{\vspace{1mm}\textbf{Q8}: Can you stop the car within the desired distance and time under the one-pedal compensated regenerative brake condition?} & 4.2 & 0.87\\ \hline
				\multicolumn{1}{m{5.75in}}{\vspace{1mm}\textbf{Q9}: Does the compensated brake pedal offer a conventional brake feel? } & 4.3 & 0.64 \\ \hline
				\multicolumn{3}{m{6.75in}}{\vspace{1mm}\textbf{} \hspace{152mm} Frequency}  \\ \hline
				\multicolumn{1}{m{5.75in}}{\vspace{1mm}\textbf{Q10}: Do you prefer the one-pedal driving or the two-pedal driving condition?} & 50\%  & 50\% \\ \hline
				\multicolumn{1}{m{5.75in}}{\vspace{1mm}\textbf{Q11}: Do you prefer the compensated or the uncompensated regenerative braking condition?} & 90\%  & 10\% \\ \hline
			\end{tabular}}\vspace{-\baselineskip}
		\end{table*}	
	\end{center}

\vspace{-3\baselineskip}
\subsection{Qualitative Metrics}

Survey questions together with their summary statistics are presented in Table~\ref{Tab:survey}.
The Cronbach's $\alpha$ for the questionnaire is evaluated to be greater than $0.96$, indicating a high reliability of the survey.

\vspace{-.5\baselineskip}
\section{Discussion and Conclusion}
\label{sec:Discussion and Conclusion}


Safety is one the key aspect for evaluating the driving performance. The number of hard brakings is a commonly used safety metric, as it is important for the drivers to be able to predict the stopping distance and safely decelerate the vehicle accordingly. The addition of regenerative braking results in a nonintuitive brake pedal force to deceleration mapping that significantly reduces the driver performance in terms of the need for hard brakings. Given that the regenerative braking is highly nonlinear and strongly affected by the instantaneous state of the vehicle, long training periods may be necessary for drivers to adjust to this nonintuitive brake mapping. Compensation of haptic pedal feel recovers the natural brake pedal feel by removing the nonlinearities and the strong dependence to the instantaneous state. In the compensated case, there exists a linear mapping between the pedal force and the total braking force that results in a significant decrease in the need for hard brakings, for both one and two pedal driving conditions.

In terms of the number of hard brakings, compensation has a larger positive effect for the two-pedal cooperative braking. While in the compensated case, both one pedal and two-pedal case have similar performance, the performance of two-pedal cooperative braking is significantly worse for the uncompensated case, as the brake pedal stays very soft during the regenerative braking phase of two-pedal cooperative braking and then suddenly stiffens, causing the drivers to overshoot the proper pedal position. On the other hand, drivers mostly experience the reaction forces from the friction brake during one-pedal driving, which results in a relatively more predictable pedal behaviour, even though the deceleration mapping is still nonlinear.

According to the survey results, the volunteers strongly agree that there is a significant intervention in  the two-pedal uncompensated braking condition, while they strongly disagree with the existence of intervention  in the compensated case. The volunteers also disagree that one-pedal compensated and uncompensated conditions have intervention. This result is also attributed to the relatively more predictable pedal forces in the uncompensated one-pedal driving case.

Consequently, for safe driving, compensated regenerative braking conditions is strongly preferred by 90\% of the volunteers and quantitatively advantageous, especially in two-pedal cooperative braking. The beneficial effect of compensation is comparatively smaller in one pedal driving, while the difference is still statistically significant and existence of compensation results in substantial quantitative effect size in terms of the number of hard brakings. Hence, haptic pedal feel compensation is highly recommended for both driving conditions  to enable more predictable decelerations of the vehicle.

In terms of the driving performance, the volunteers were able to adequately adjust the distance between two vehicles in all conditions with no significant differences. In the survey, the volunteers agree that they can stop the car within the desired distance in both compensated conditions, while they are neutral to both  uncompensated conditions. However, hard brakings negatively affect driving, as  sharp decelerations are disturbing. Consequently, for the driving performance, compensated regenerative braking conditions are both more strongly preferred and advantageous.

In terms of the throttle use, one pedal driving necessitates significantly more use of the accelerator, as the use of throttle is required even for coasting. In terms of the total regenerated brake energy, one-pedal driving results in a significantly higher regeneration level, since regenerative brakes are more frequently used, as this type of brake engages as soon as the driver releases the throttle pedal.  The compensation does not have a significant effect on throttle use or the total regenerated brake energy, as the need for cooperative braking is quite infrequent compared to throttle use and mild regenerative braking during the simulated pursuit tracking task.

While one-pedal driving recovers significantly more more energy from regeneration, this does not necessarily imply better energy efficiency of the vehicle, as it also results in significantly more throttle use. Proper evaluation of the overall energy efficiency requires further investigation, as a more detailed dynamic model of the vehicle and efficiency of the power electronics during acceleration and regeneration need to be considered.


In conclusion, compensation of haptic pedal feel has been shown to be advantageous, especially in term of safety and driver preferences, for both two-pedal cooperative braking and one-pedal driving. While the volunteers equally prefer and can effectively utilize both two-pedal and one-pedal driving conditions, the beneficial effects of haptic pedal feel compensation is shown to be larger for the two-pedal cooperative braking case.


\vspace{-3mm}
\bibliographystyle{ieeetran}

\end{document}